\documentclass[5p,times,procedia]{elsarticle}
\usepackage{subcaption}

\usepackage{placeins}
\usepackage{float}
\usepackage{multirow}

\usepackage{amsmath}
\usepackage{amssymb}
\usepackage{enumitem}

\usepackage{ecrc}

\volume{00}

\firstpage{1}

\journalname{Procedia CIRP}

\runauth{Christoph Huber et al.}

\jid{trpro}

\usepackage{amssymb}

\usepackage[figuresright]{rotating}

\usepackage[bookmarks=false]{hyperref}
    \hypersetup{colorlinks,
      linkcolor=blue,
      citecolor=blue,
      urlcolor=blue}

\usepackage[capitalize]{cleveref}
\crefname{section}{Sec.}{Secs.}
\Crefname{section}{Section}{Sections}
\Crefname{table}{Table}{Tables}
\crefname{table}{Tab.}{Tabs.}

\begin{document}
\begin{frontmatter}



\dochead{Proceedings of the 58th CIRP Conference on Manufacturing Systems 2025}%

\title{Fully-Synthetic Training for Visual Quality Inspection in Automotive Production}


\author[]{Christoph Huber$^{\text{a,b,*}}$\corref{*}} 
\author[b]{Dino Knoll}
\author[a]{Michael Guthe}

\address[a]{University of Bayreuth, Universitätsstraße 30, 95447 Bayreuth, Germany}
\address[b]{BMW Group, Petuelring 130, 80809 Munich, Germany}

\aucores{* Corresponding author. Tel.: +49-151-601-68945. {\it E-mail address:} christoph.ch.huber@bmw.de}

\begin{abstract}
  Visual Quality Inspection plays a crucial role in modern manufacturing environments as it ensures customer safety and satisfaction. The introduction of Computer Vision (CV) has revolutionized visual quality inspection by improving the accuracy and efficiency of defect detection. However, traditional CV models heavily rely on extensive datasets for training, which can be costly, time-consuming, and error-prone. To overcome these challenges, synthetic images have emerged as a promising alternative. They offer a cost-effective solution with automatically generated labels. In this paper, we propose a pipeline for generating synthetic images using domain randomization. We evaluate our approach in three real inspection scenarios and demonstrate that an object detection model trained solely on synthetic data can outperform models trained on real images.
\end{abstract}


\end{frontmatter}



\section{Introduction}
\label{sec:intro}

Artificial Intelligence (AI) and CV are critical components in today's manufacturing plants \cite{charan_future_2022, schuh_data_2019, zhou_computer_2023}. In particular, visual quality inspection — a vital process for ensuring production quality — benefits significantly from deep learning-based CV applications. These applications offer consistent and precise defect detection, operate at higher speeds, and reduce labor costs compared to traditional human inspectors \cite{ebayyeh_review_2020, konstantinidis_role_2021}.

However, training such models requires substantial amounts of labeled training data in order to achieve reliable performance \cite{tobin_domain_2017-1}. 
Generating datasets of this scale in a manufacturing environment presents significant challenges. Certain production variants occur infrequently, making it difficult to collect a sufficient number of images for these cases. Similarly, many types of defects are rare, and intentionally creating these defects for data collection can significantly increase development costs \cite{jain_synthetic_2022}.
Moreover, changes to the production system can degrade the performance of existing inspection models, requiring regular retraining and additional data collection \cite{moonen_cad2render_2023}.

\begin{figure}
  \centering
  \begin{subfigure}{0.4\linewidth}
    \centering
    \includegraphics[width=.85\linewidth]{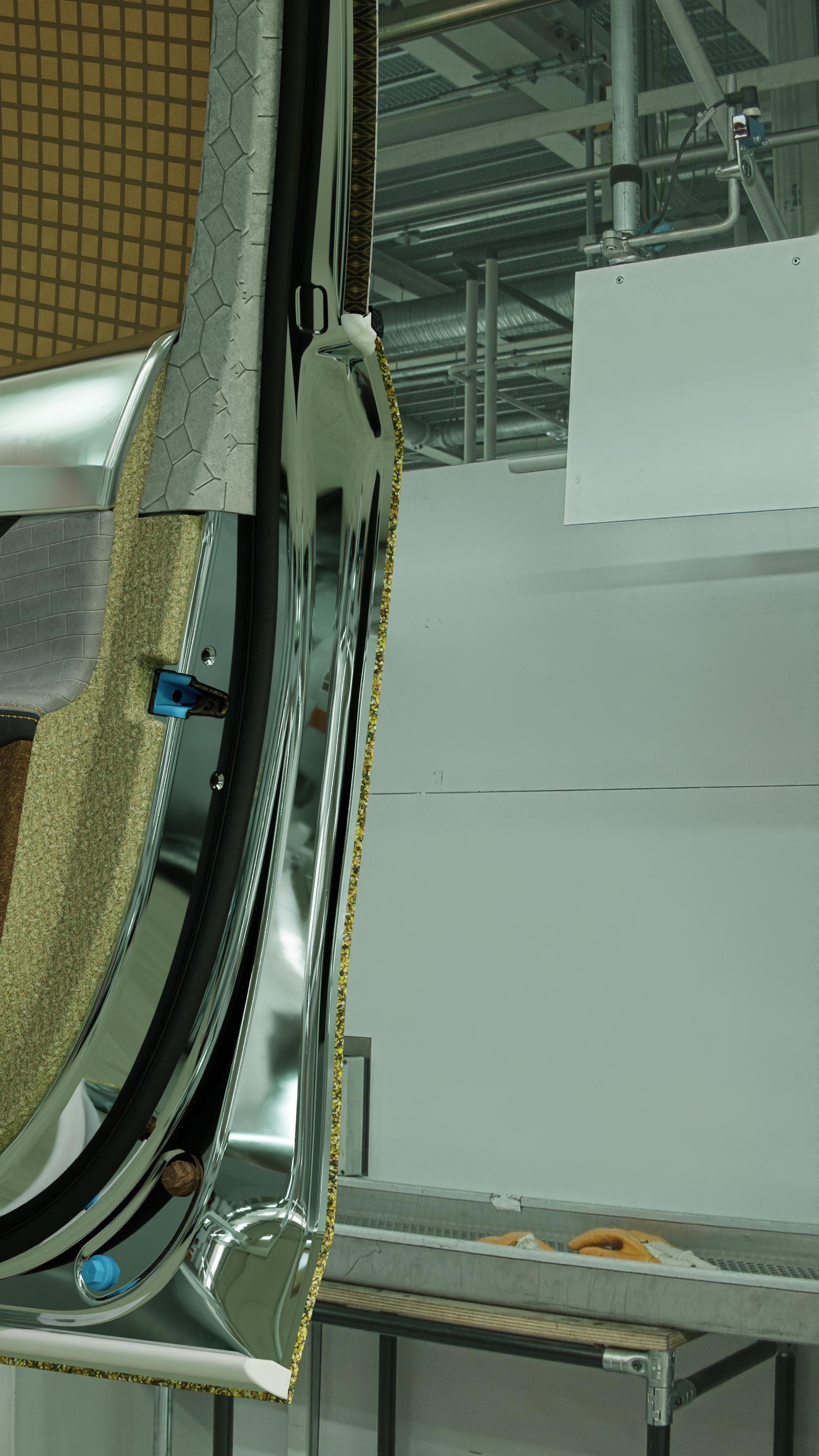}
  \end{subfigure}
  \hspace{0.067\linewidth}
  \begin{subfigure}{0.4\linewidth}
    \centering
    \includegraphics[width=.85\linewidth]{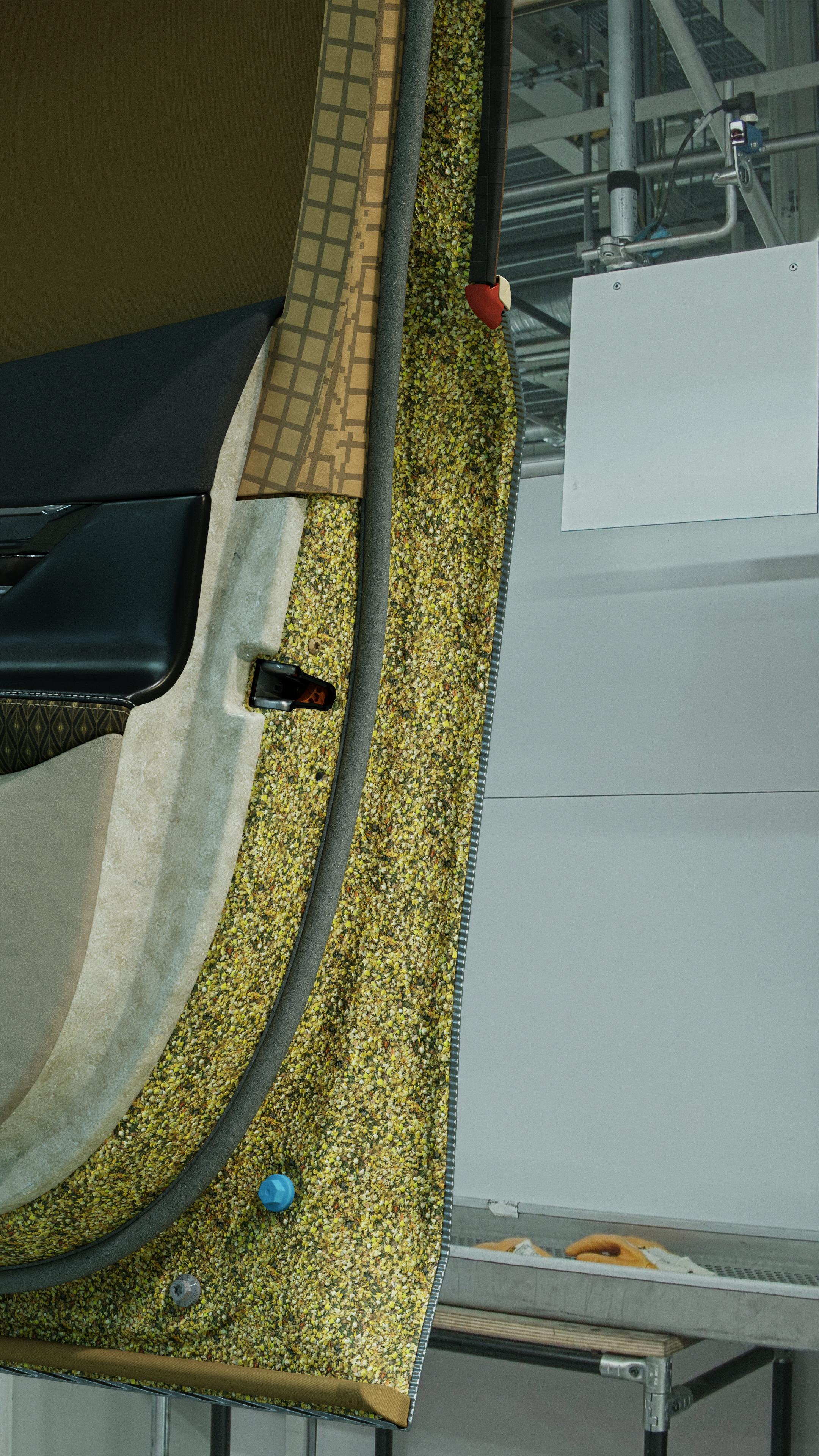}
  \end{subfigure}
   \caption{
    Some examples of our synthetic images for the inspection shown in \Cref{fig:tvm_lock}. They illustrate the variety of materials utilized during image synthesis.
    }
   \label{fig:test_imgs}
\end{figure}

Synthetic data offers a compelling alternative to real-world datasets because it reduces the effort required to collect images and annotations \cite{hinterstoisser_annotation_2019-1,jain_synthetic_2022,manettas_synthetic_2021,mayershofer_towards_2021}.
Synthetic data allows developers to artificially generate massive datasets with perfect labels, providing precise control over the content and distribution of the dataset. 
However, the use of such datasets comes with its own challenges, namely, how to bridge the domain gap \cite{tobin_domain_2017-1}. 
The domain gap describes the characteristic differences between real and synthetic images, which can often deteriorate detection performance in real-world inspection tasks, as the trained model fails to generalize across domains \cite{mayershofer_towards_2021}.

To address the domain gap, a technique known as domain randomization can be employed. First introduced by Tobin et al. \cite{tobin_domain_2017-1}, it adds extensive variance into the synthesis process to help the CV model become more robust to real data. The underlying concept is that by significantly randomizing the synthetic training data, the model will "perceive" reality as just another variation within the same domain \cite{hinterstoisser_annotation_2019-1}.

In this paper, we investigate the use of synthetic data for training an object detection model, aiming to deploy it for visual quality inspection in a vehicle assembly line. Typical inspections in this field focus on detecting the presence of relevant components and, when multiple options exist due to country regulations or optional extras, identifying the specific variant. Common use cases include verifying mounting hardware such as screws and clamps, detecting interior and exterior options, and ensuring the presence of easily overlooked components. 

Most relevant studies in the field \cite{eversberg_generating_2021, grassler_creating_2024, horvath_object_2023, mayershofer_towards_2021, zhu_towards_2023} concentrate on applications that involve identifying a small number of large, distinctive components against plain backgrounds. While these studies demonstrate promising results, they do not adequately reflect the requirements of an automotive production line, where the relevant components are typically smaller and integrated into larger assemblies. Additionally, many of these studies still suffer from the domain gap, leading to models that are too unreliable for real-world inspections. To mitigate this issue, these works often incorporate some real data during training to enhance performance.

We demonstrate through three real-world inspections that a model trained exclusively on synthetic data, incorporating domain randomization techniques, can match and, in some cases, even surpass the performance of a benchmark model trained on a limited set of real images. Additionally, we conduct an extensive ablation study on one use case to explore the impact of various factors on the model's ability to generalize across domains. 
Some examples of the synthetic data used in this study can be seen in \Cref{fig:test_imgs}.

To the best of our knowledge, this is the first study to show that domain randomization allows the application of object detection models trained purely on synthetic data for visual quality inspection in a production setting. Our research paves the way for further investigations into a wider range of applications. In the remainder of this paper, we will first discuss related work in the field. We will then proceed with a description of our state-of-the-art pipeline for generating synthetic images, followed by a demonstration of the use cases on which our research is based. Finally, we will present the details of our experiments and draw conclusions based on our findings.

\section{Related Work}
\label{sec:related}

Researchers have introduced a wide range of methods for utilizing synthetic data in CV. 
One approach attempts to generate photo-realistic images, that match the real world as closely as possible \cite{akar_synthetic_2022, de_roovere_sim--real_2024, manettas_synthetic_2021, moonen_cad2render_2023}.
These methods have been shown to generalize well between domains, but producing high-quality images is a significant challenge due to the complexity of accurately modeling lighting conditions and surface textures.
This makes it difficult to use photo-realistic data in production, as the effort for generating synthetic images can quickly outweigh the effort for producing real images.

Other works use Generative Adversarial Networks (GANs) \cite{goodfellow_generative_2014} to alter the synthetic images in such a way that they become indistinguishable from real images \cite{hu_synthetic_2023, rozanec_synthetic_2023, sixt_rendergan_2018}. These models iteratively optimize a generator and a discriminator network, where the generator learns to transform source images into the target domain, while the discriminator aims to differentiate between real images and the generator's output. 
Alternatively, Autoencoders \cite{kramer_nonlinear_1991} can be used to transform both synthetic and real images into a common representation. These models are trained to compress data into a condensed format and subsequently reconstruct the input from this compressed state. \cite{abubakr_learning_2022, ghifary_deep_2016-1} use this principle to jointly train an autoencoder on data from both the source and the target domain. The compressed representation at the center of the autoencoder can then be used as a domain-invariant input for the classification network. 
While both GANs and autoencoders can help mitigate the impact of the domain gap, they still require real data during training. 

In this paper, we employ Domain Randomization \cite{tobin_domain_2017-1}, a technique that modifies elements of the rendered scene to produce non-realistic images with diverse variations. The core idea is that by introducing substantial variability into the training data, real-world data can be perceived as just another variation within the same domain \cite{hinterstoisser_annotation_2019-1}. Research has shown that models trained using domain randomization can generalize effectively to real-world data without requiring actual training examples \cite{eversberg_generating_2021, hinterstoisser_annotation_2019-1, tobin_domain_2017-1}.
Previous studies have indicated that synthetic data can be beneficial for industrial CV applications \cite{grassler_creating_2024, horvath_object_2023, mayershofer_towards_2021, zhu_towards_2023}. However, these works are typically limited to simple inspections involving easily distinguishable objects and often rely on additional fine-tuning with real images to achieve reliable performance.




\section{Image Generation Pipeline}
\label{sec:pipeline}

In this section, we discuss our pipeline for generating synthetic training data. To ensure that the model trained solely on synthetic images performs well on real data, we apply domain randomization \cite{tobin_domain_2017-1} to introduce extensive variance into the training images during rendering.

For synthetic data to be effectively utilized in a manufacturing setting and to address the challenges of data gathering, it must meet certain requirements. It should encompass all possible production variants, account for all potential defects, and the 3D model must accurately represent the real geometry, reflecting the same state of the assembled product as seen in real images.
The image generation pipeline must automatically generate labels to minimize manual effort. The materials used for domain randomization must be diverse in structure and color, and the High Dynamic Range Images (HDRIs) used for lighting should mimic the characteristics of the real scene. 

The rendering process can be divided into three steps: First, the required CAD data is collected and imported into a rendering software. In this work, we utilize NVIDIA Omniverse \cite{omniverse} for its accurate lighting and user-friendly Python API. Next, the real-world scene is recreated by positioning a virtual camera in approximately the same location as in reality. A primitive cube is then placed behind the scene, onto which a real background image is projected. Finally, we use Omniverse's Python API to automatically apply the following randomizations to each image, as derived from Eversberg and Lambrecht \cite{eversberg_generating_2021}:

We first randomly sample a material from a collection of 115 diverse options for each component. For lighting, we utilize 16 indoor HDRIs. To further enhance variability, we rotate the HDRI around its vertical axis and modify the color and intensity of the emitted light. 
The camera's position and rotation are adjusted around the initial orientation within limits that ensure the region of interest remains visible in the images. Finally, Gaussian noise is added to simulate the effects of sensor noise.

\begin{figure}
  \centering
  \includegraphics[width=.9\linewidth]{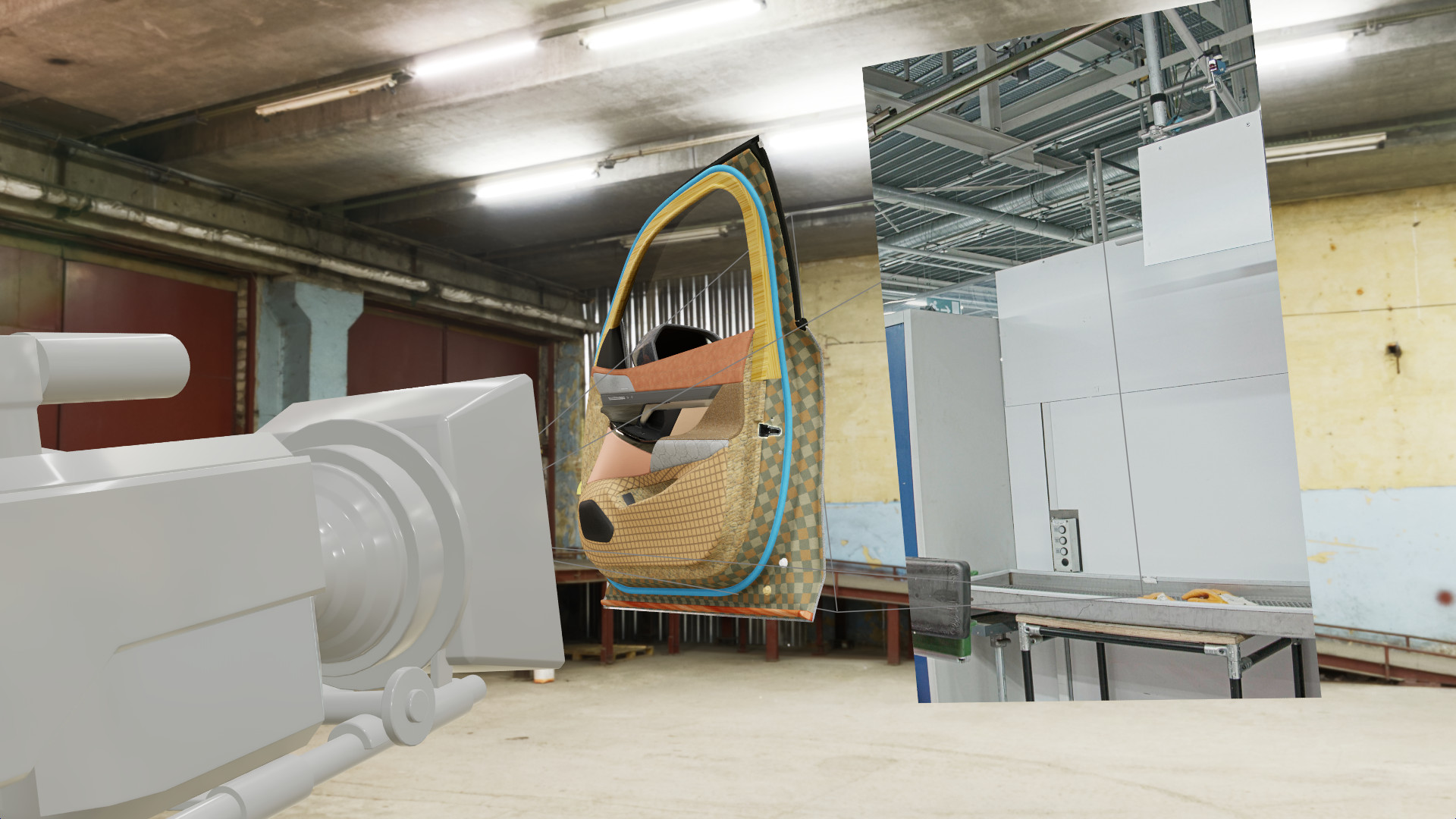}
   \caption{Illustration of a recreated 3D scene within NVIDIA Omniverse, based on the real scene shown in Figure \ref{fig:tvm_lock}. 
   }
   \label{fig:scene}
\end{figure}

We employ path tracing as the image formation algorithm. Mayershofer et al. \cite{mayershofer_towards_2021} demonstrated that physically based renderings like path tracing offer higher information content per image than simpler methods, enabling more effective model training with fewer images.
With this pipeline, images with a resolution of 1920x1080 are generated in approximately 2 seconds per frame. The structure of an exemplary scene can be seen in Figure \ref{fig:scene}.

\section{Experiments}
\label{sec:experiments}
In this section, we discuss the experiments and results of our research, which demonstrate the benefits of our image generation strategy. We begin by introducing the three use cases on which our study is based. Then, we describe the experimental setup and proceed to demonstrate how our synthetic images enable us to train object detection models without any real data.

\subsection{Test Cases}
\label{ssec:test_cases}
\begin{figure*}
  \centering
  \begin{subfigure}{0.3\linewidth}
    \centering
    \includegraphics[width=.4\linewidth]{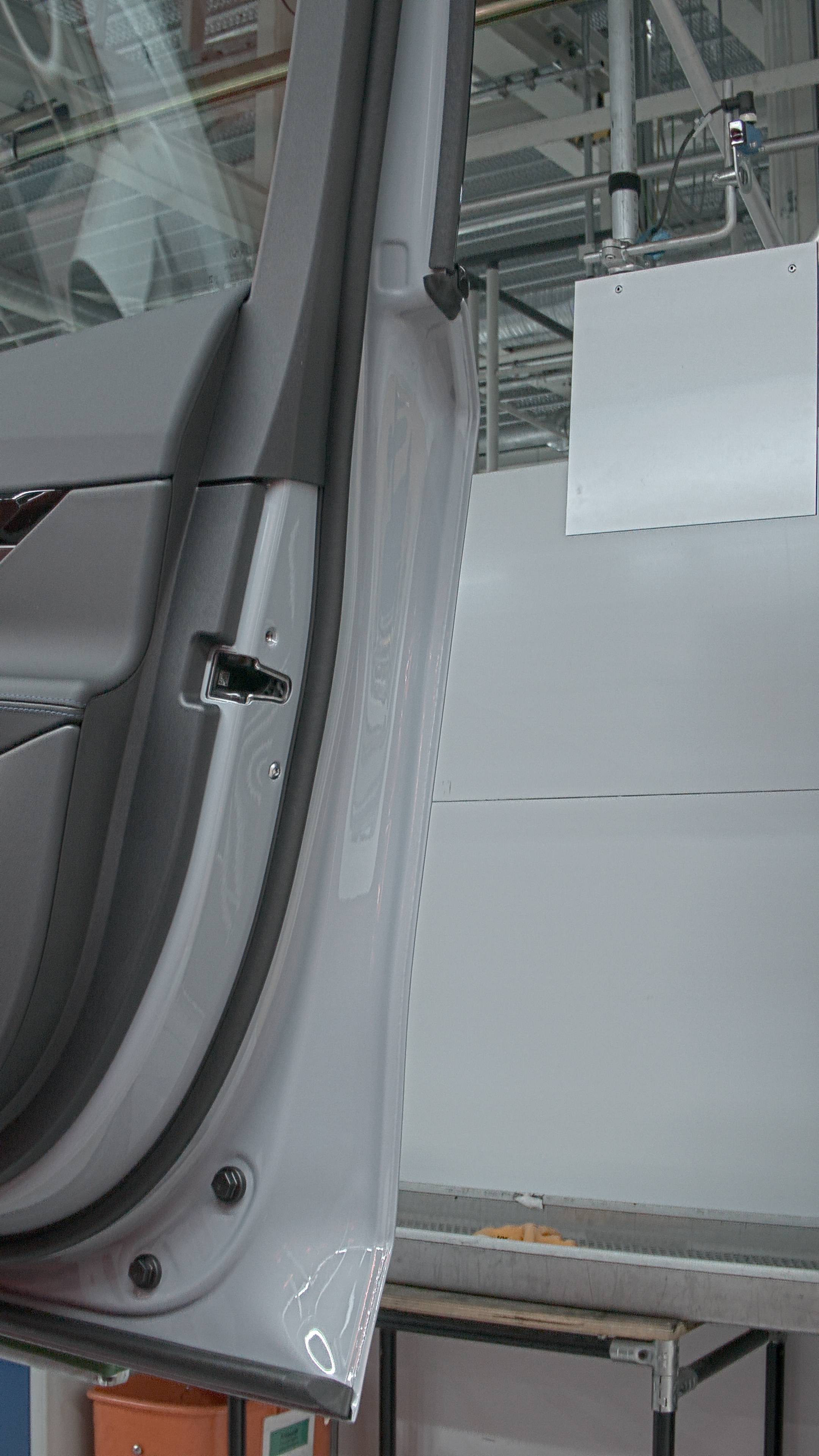}
    \caption{Door lock inspection.}
    \label{fig:tvm_lock}
  \end{subfigure}
  \hfill
  \begin{subfigure}{0.3\linewidth}
     \centering
     \includegraphics[width=.4\linewidth]{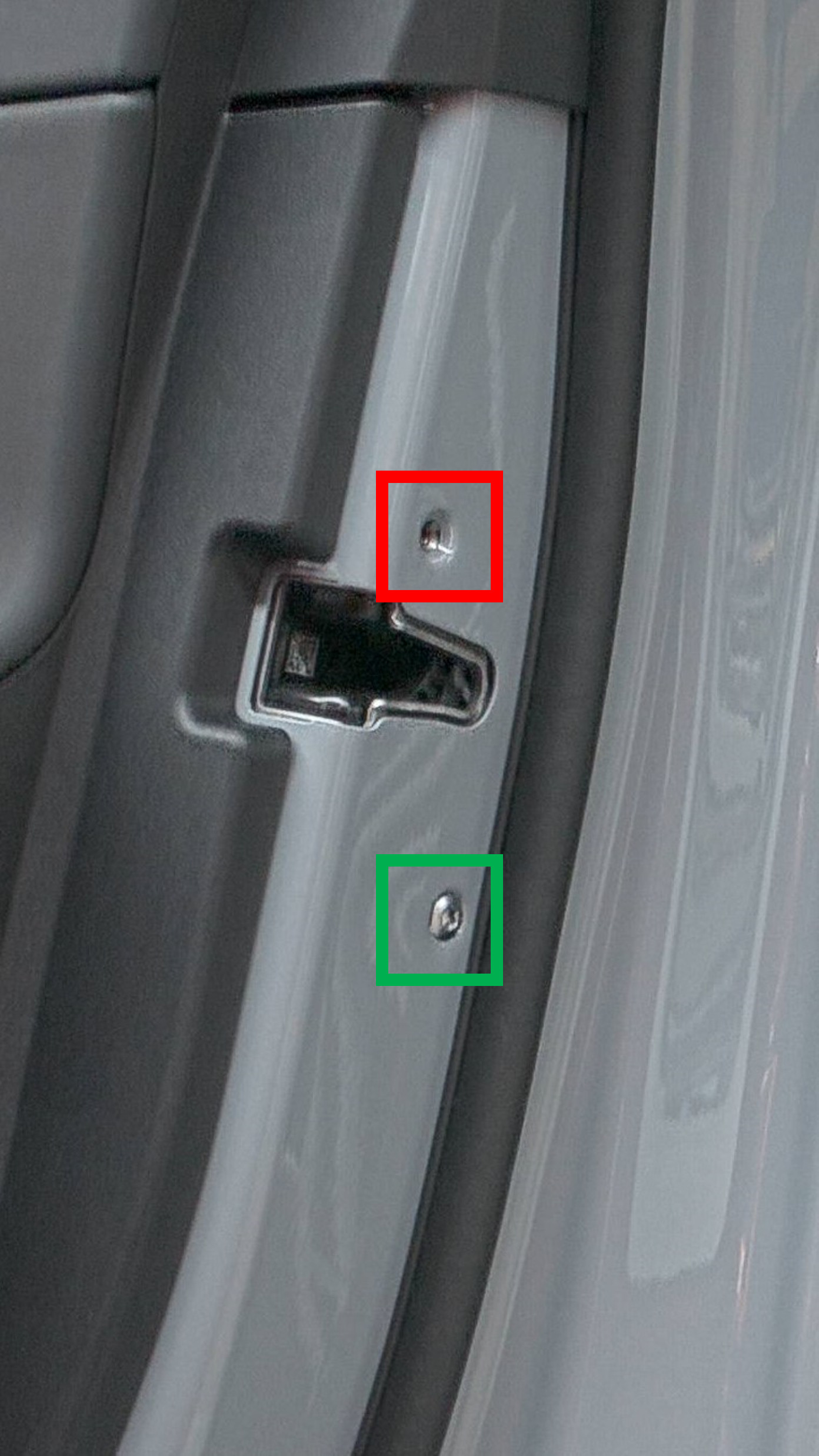}
    \caption{Example of a missing screw.}
    \label{fig:tvm_lock_crop}
  \end{subfigure}
  \hfill
  \begin{subfigure}{0.3\linewidth}
     \centering
     \includegraphics[width=.4\linewidth]{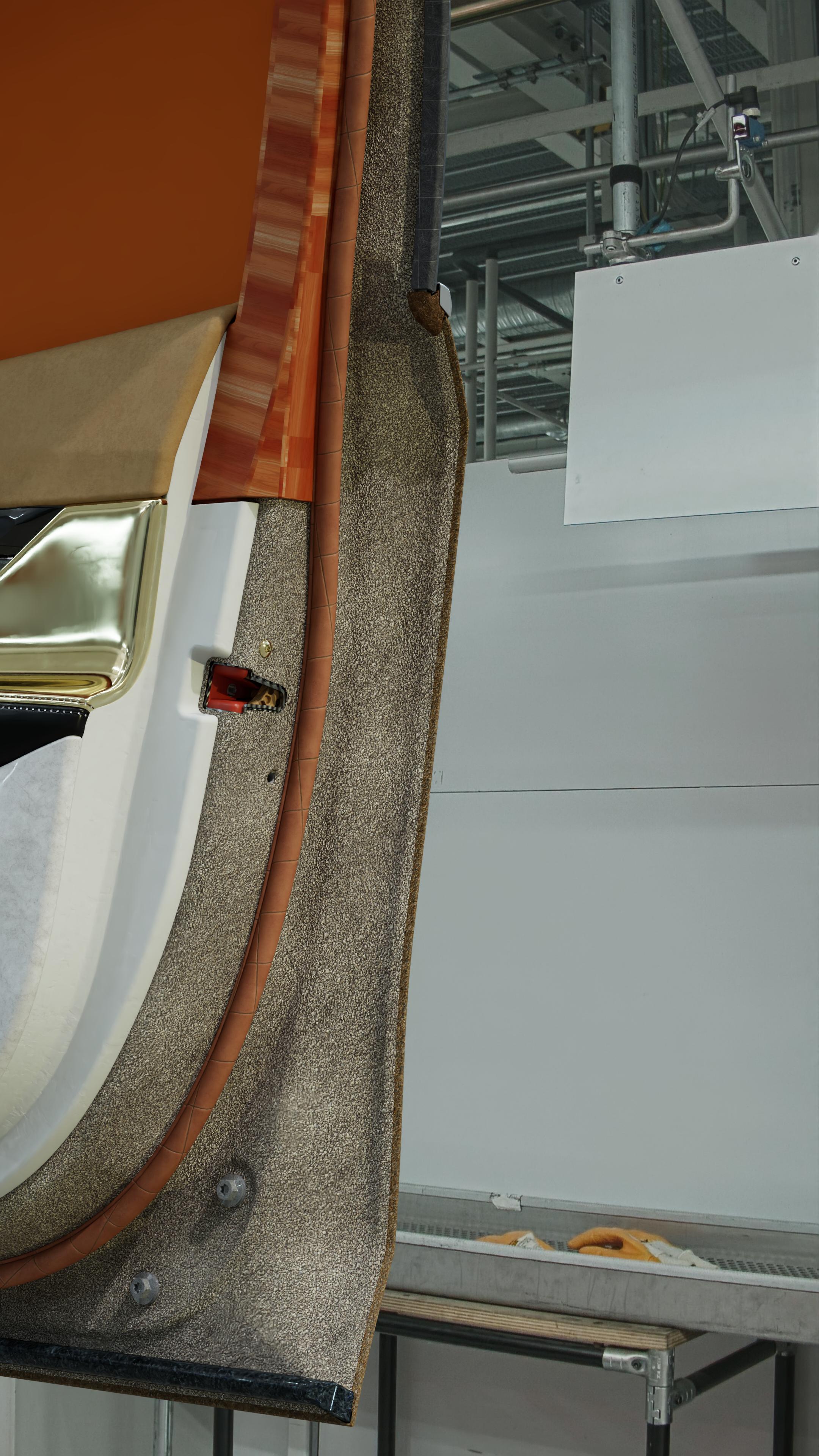}
    \caption{Synthetic image.}
    \label{fig:tvm_lock_synth}
  \end{subfigure}
  \begin{subfigure}{0.3\linewidth}
    \centering
    \includegraphics[width=.9\linewidth]{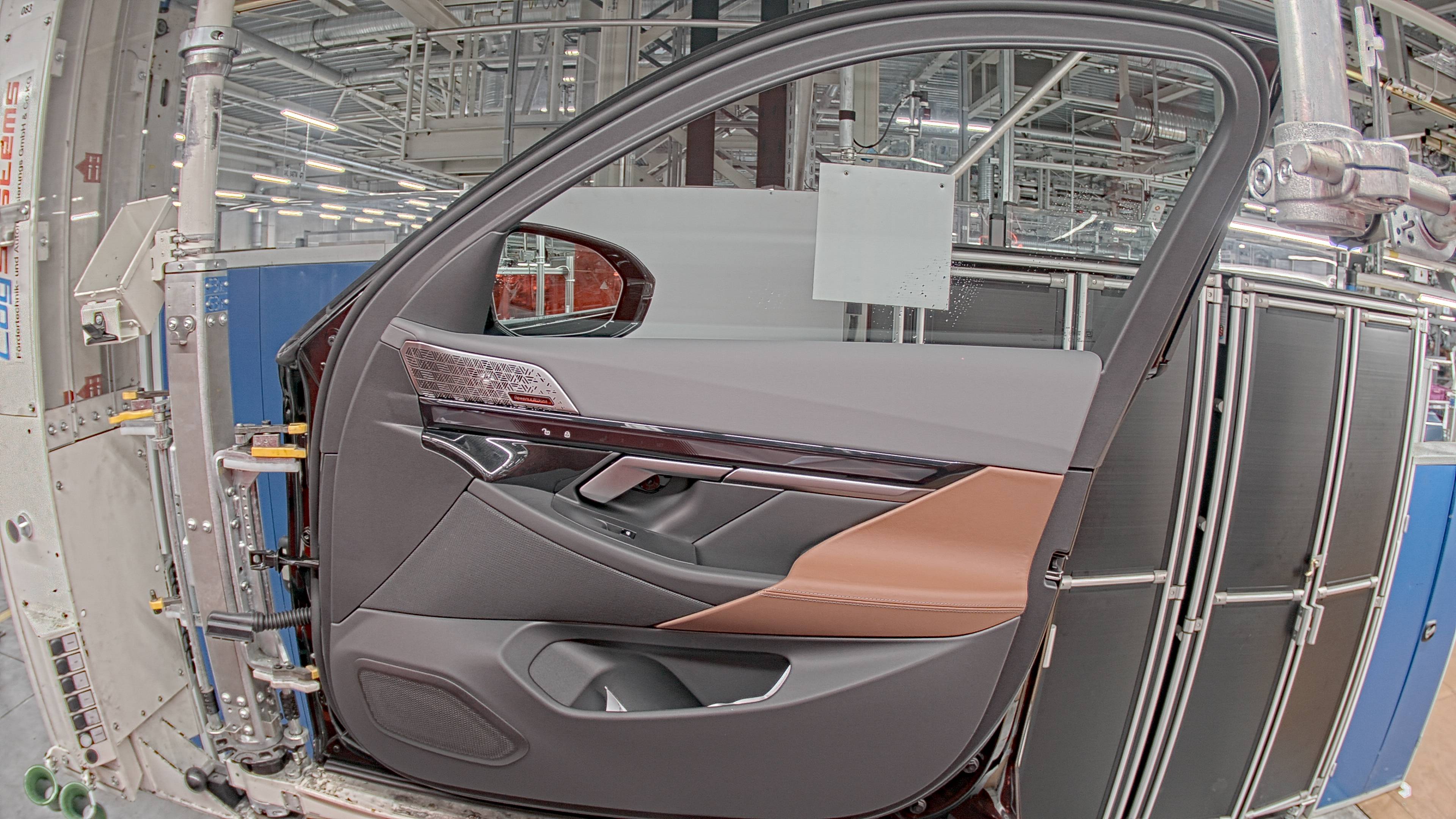}
    \caption{Door opener cover.}
    \label{fig:tvm_opener}
  \end{subfigure}
  \hfill
  \begin{subfigure}{0.3\linewidth}
    \centering
     \includegraphics[width=.9\linewidth]{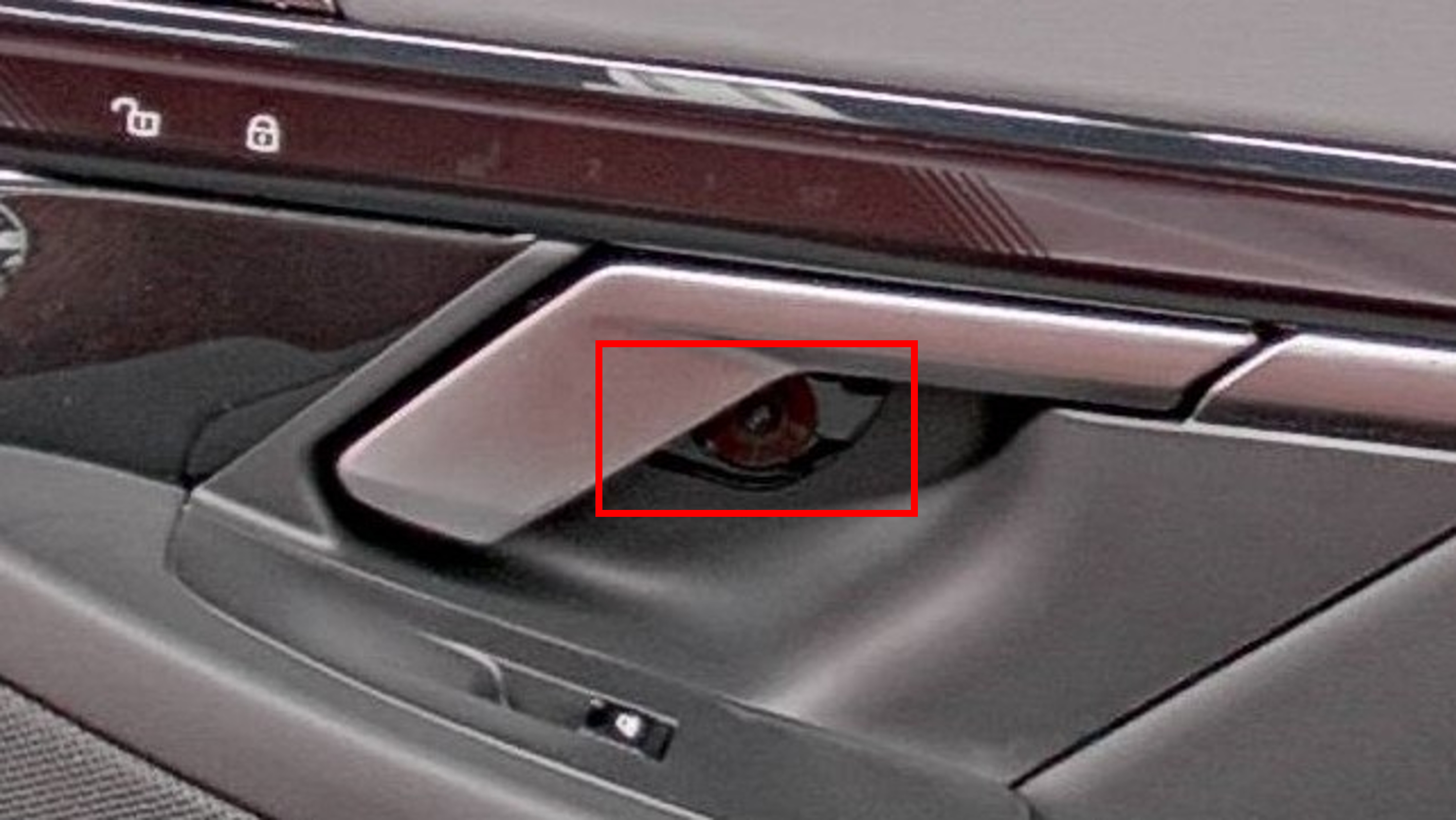}
    \caption{Example of a missing cover.}
    \label{fig:tvm_opener_crop}
  \end{subfigure}\hfill
  \begin{subfigure}{0.3\linewidth}
    \centering
     \includegraphics[width=.9\linewidth]{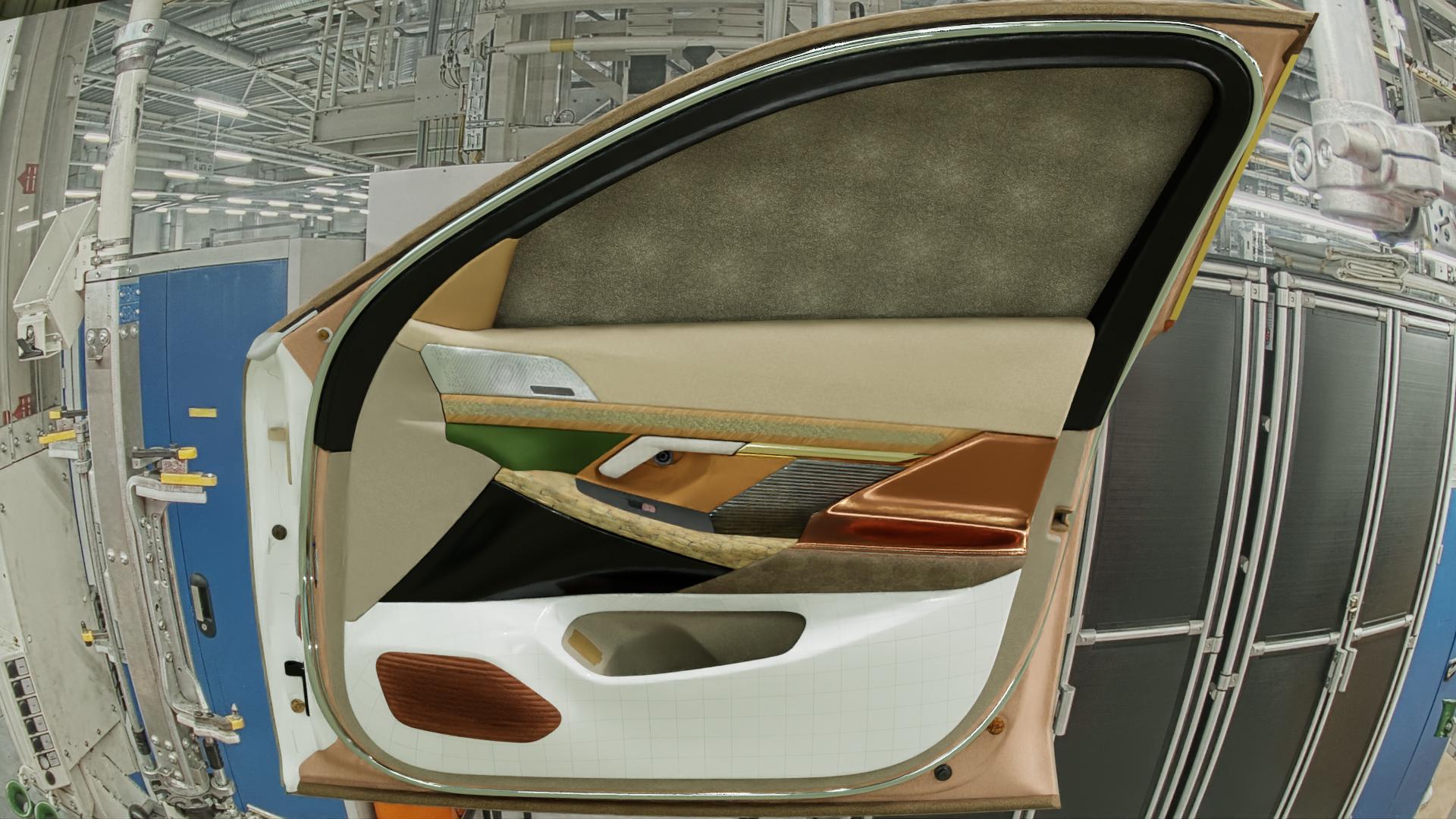}
    \caption{Synthetic image.}
    \label{fig:tvm_opener_synth}
  \end{subfigure}
  \begin{subfigure}{0.3\linewidth}
    \centering
    \includegraphics[width=.9\linewidth]{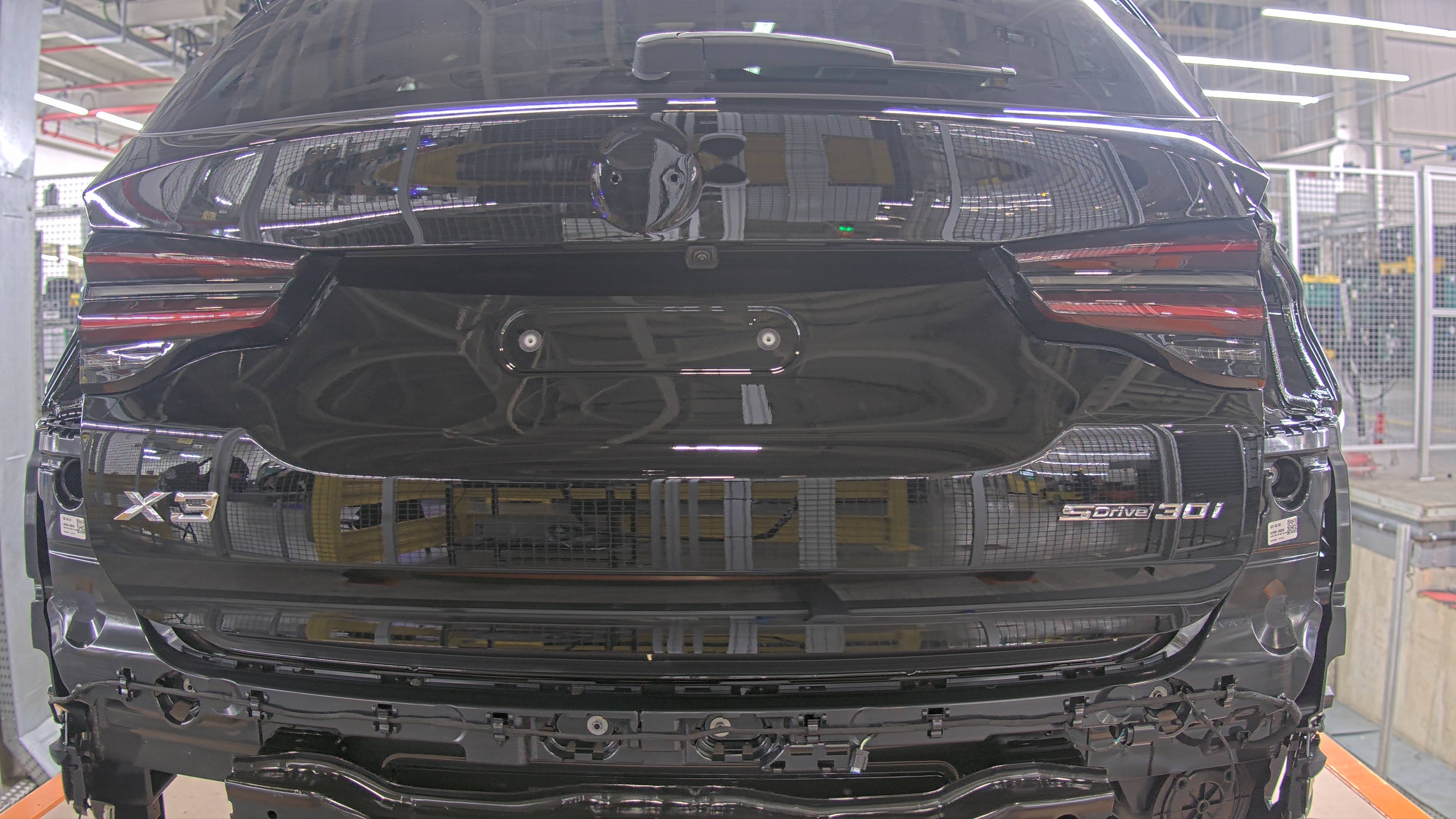}
    \caption{Type label inspection.}
    \label{fig:type_label}
  \end{subfigure}
  \hfill
  \begin{subfigure}{0.3\linewidth}
    \centering
     \includegraphics[width=.9\linewidth]{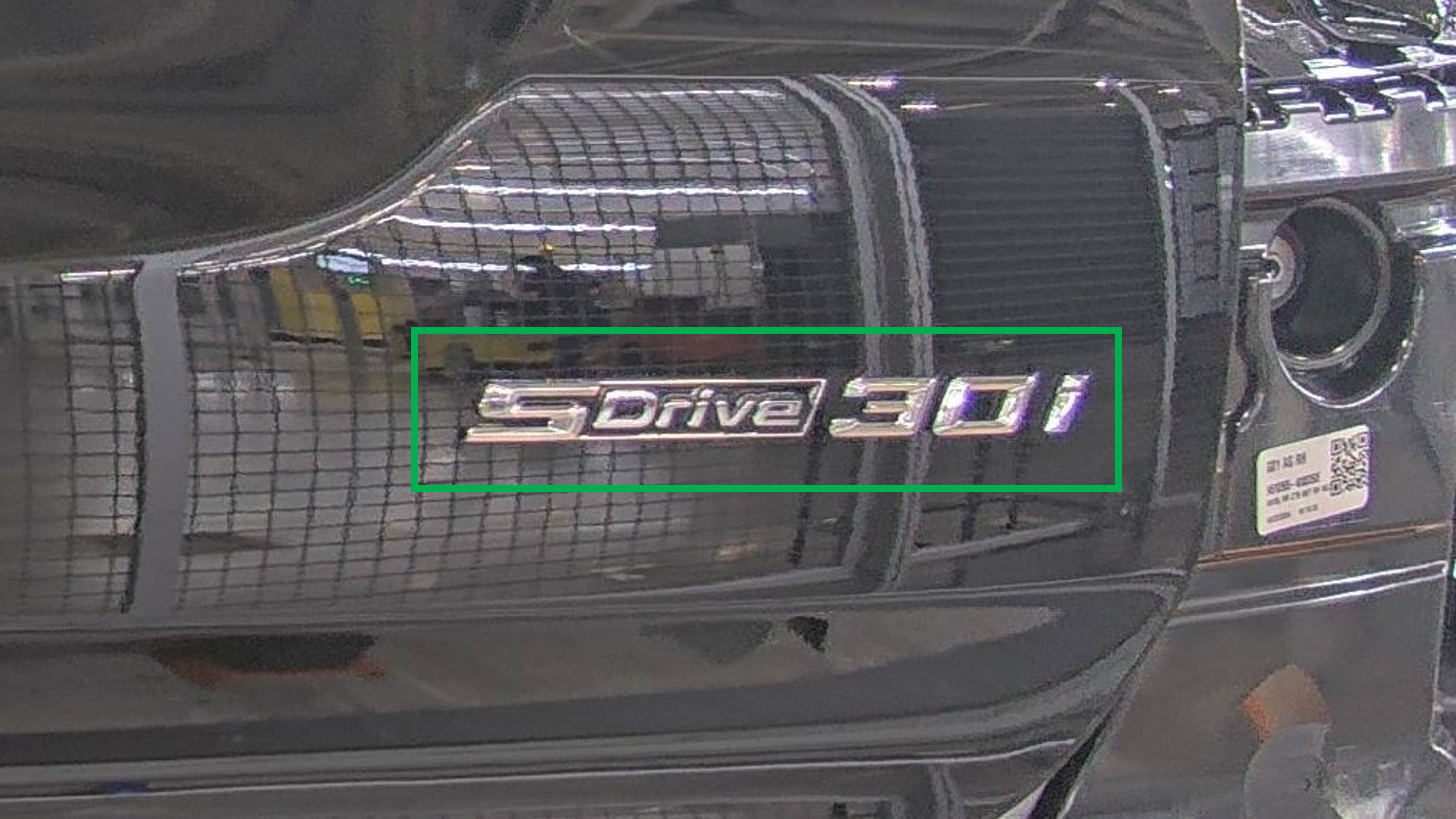}
    \caption{Example of an sDrive30i badge.}
    \label{fig:type_label_crop}
  \end{subfigure}
  \hfill
  \begin{subfigure}{0.3\linewidth}
    \centering
     \includegraphics[width=.9\linewidth]{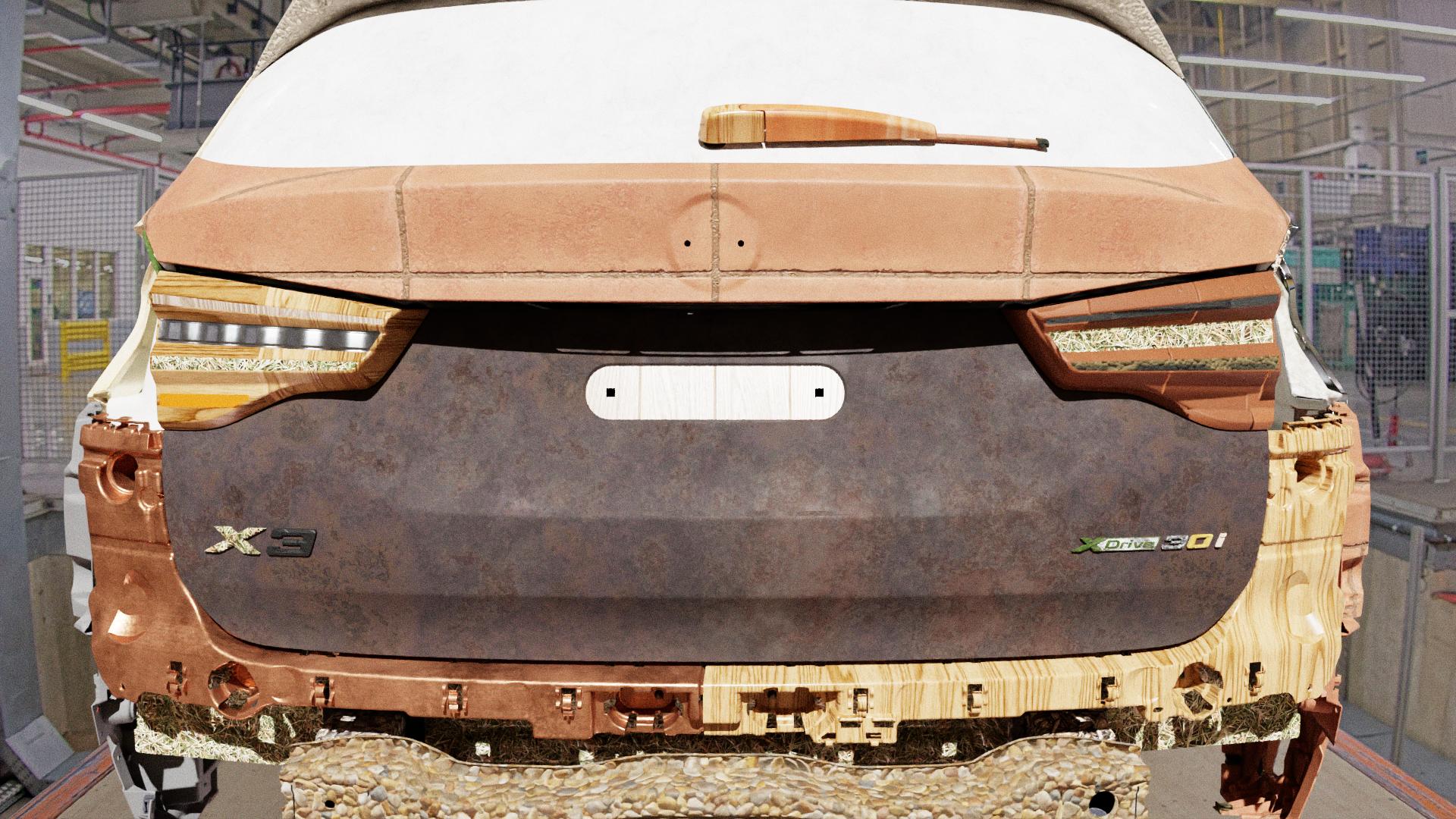}
    \caption{Synthetic image.}
    \label{fig:type_label_synth}
  \end{subfigure}
  \caption{The left column presents the three test cases we selected to validate our approach. The model is presented with the full image. The middle column shows crops of the features we want to inspect. Green boxes indicate an OK inspection, red boxes mark an NOK. The right column shows examples of synthetic images.}
  \label{fig:usecases}
\end{figure*}

We evaluate our method's applicability to quality inspection by training object detection models on three real-world inspection use cases, which are shown in Figure \ref{fig:usecases}.

\begin{enumerate}[wide, labelwidth=!, labelindent=0pt]
    \item \textbf{Door Lock Inspection:} Here we want to ensure the presence of two screws located above and below the door lock. This test is representative for other use cases, in which the presence of safety-critical mounting parts, such as screws, clips, and clamps, must be confirmed.
    \item \textbf{Door Opener Cover:} This use case checks for a plastic cover under the door handle. While similar to the first case, it is more challenging due to the object being partially obscured, leading to poor illumination. Additionally, the cover and its surroundings are made of the same material, making it difficult to identify object boundaries. 
    \item \textbf{Type Label Inspection:} This use case inspects the type labels on the rear of the car, featuring 12 different labels that present a new challenge compared to the first two cases. The model must distinguish between a larger number of similar options, such as xDrive20d, xDrive20i, and sDrive20i.
\end{enumerate}

For all use cases, the model is presented with the full image during inference, since the region of interest can shift from image to image.
The cameras capture images in 4K resolution, but they are down-sampled to 1024 pixels for inference.

\subsection{Experiment Settings}
For our experiments, we selected YOLOv8s \cite{jocher_ultralytics_2023} and Faster R-CNN with a ResNet50 backbone \cite{RenHG015} as state-of-the-art object detection architectures. Both models were initialized with weights pre-trained on the COCO dataset \cite{fleet_microsoft_2014}. They were trained for 200 epochs using an SGD optimizer, an initial learning rate of 0.1, an input size of 1024, and a batch size of 16. Each training took approximately four hours utilizing two NVIDIA A6000 GPUs.
We analyze the model's performance using the Accuracy (A), Precision (P), Recall (R), and mean Average Precision at the 50\% Intersection of Union (mAP).

\subsection{Experimental Evaluation}
\label{ssec:poc}
Here we show that our data generation process allows training a model purely on synthetic data that performs as well as or better than a traditionally trained model on common inspection tasks. For each use case presented in \Cref{ssec:test_cases}, we train two instances: one on synthetic data and the other as a benchmark on real images. Each synthetic dataset contains 10,000 images with an 80:20 train-validation split. The number of images in the benchmark and test datasets can be found in Table \ref{tab:ds_size}.

\begin{table}
\caption{Distribution of images in the benchmark and test datasets. For use cases 1 and 2, the table displays the count of acceptable (OK) versus non-acceptable (NOK) images. In use case 3, each class is represented equally.}
\label{tab:ds_size}
\begin{tabular*}{\hsize}{@{\extracolsep{\fill}}lccc@{}}
    \toprule
    & Train & Val & Test \\
    \colrule
    Door Lock Screws  & 360/40   & 80/20 & 460/40  \\ 
    Door Opener Cover & 360/40   & 80/20 & 80/20  \\
    Type Label        & 1800  & 420 & 180  \\ 
    \botrule
\end{tabular*}
\end{table}

Table \ref{table:poc} presents the performance metrics when testing all instances on real images. The models trained on synthetic data achieved 100\% accuracy in two use cases, while matching or surpassing the performance of benchmark models in all experiments. This demonstrates that synthetic data can be a viable alternative to using real images for training.

\begin{table}
    \caption{Results for the three test cases presented in \Cref{ssec:test_cases}.}
\label{table:poc}
    \begin{tabular*}{\hsize}{@{\extracolsep{\fill}}lccccccccc@{}}
        \toprule
        & Model & Dataset & A & P & R & mAP \\
        \colrule
        \multirow{4}{*}{Door Lock} & \multirow{2}{*}{YOLOv8} & Benchmark & 0.99 & 0.98 & 0.98 & 0.98 \\
        & & Synthetic & 1.00 & 0.99 & 1.00 & 0.99 \\
        & \multirow{2}{*}{FRCNN} & Benchmark & 0.98 & 0.98 & 0.96 & 0.95 \\
        & & Synthetic & 1.00 & 1.00 & 1.00 & 0.99 \\
        \multirow{4}{*}{Door Opener} & \multirow{2}{*}{YOLOv8} & Benchmark & 1.00 & 1.00 & 1.00 & 0.99 \\
        & & Synthetic & 1.00 & 1.00 & 1.00 & 0.99 \\
        & \multirow{2}{*}{FRCNN} & Benchmark & 0.98 & 0.96 & 0.99 & 0.97 \\
        & & Synthetic & 1.00 & 1.00 & 1.00 & 0.99 \\
        \multirow{4}{*}{Type Label} & \multirow{2}{*}{YOLOv8} & Benchmark & 0.99 & 0.99 & 0.91 & 0.97 \\
        & & Synthetic & 0.99 & 0.98 & 0.99 & 0.99 \\
        & \multirow{2}{*}{FRCNN} & Benchmark & 0.96 & 0.96 & 0.90 & 0.95 \\
        & & Synthetic & 0.98 & 0.96 & 0.97 & 0.96 \\
        \botrule
    \end{tabular*}
\end{table}

\section{Ablation Study}
In the following experiments, we analyze the impact of various factors on a neural network's ability to transfer knowledge across the domain gap. These factors include the kind of materials assigned to the model, the scene's background, and the use of noise and distractors. 
The basis for this study is the door lock inspection presented in Figure \ref{fig:tvm_lock}. In this section, we use YOLOv8 for all experiments. 
To gain a more detailed understanding of these influences, we created a challenging new dataset with more variance. 
For each experiment, we conducted four training runs to account for training-to-training variances, and we present the average values for each metric.

\subsection{Test Data}
\label{sssec:ablation_data}

\begin{figure}
  \centering
  \begin{subfigure}{0.4\linewidth}
    \centering
    \includegraphics[width=.8\linewidth]{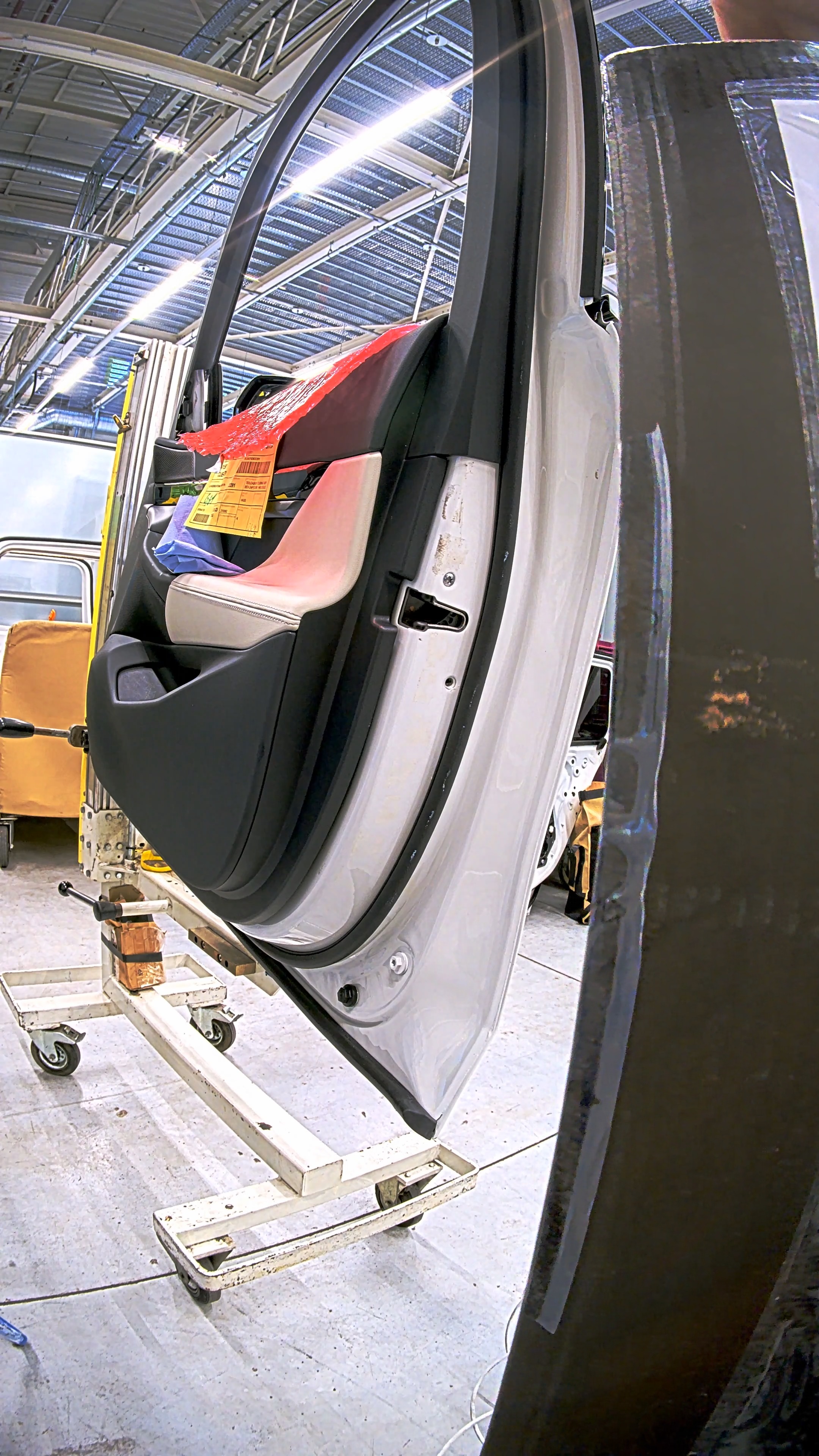}
  \end{subfigure}
  \hspace{0.067\linewidth}
  \begin{subfigure}{0.4\linewidth}
    \centering
    \includegraphics[width=.8\linewidth]{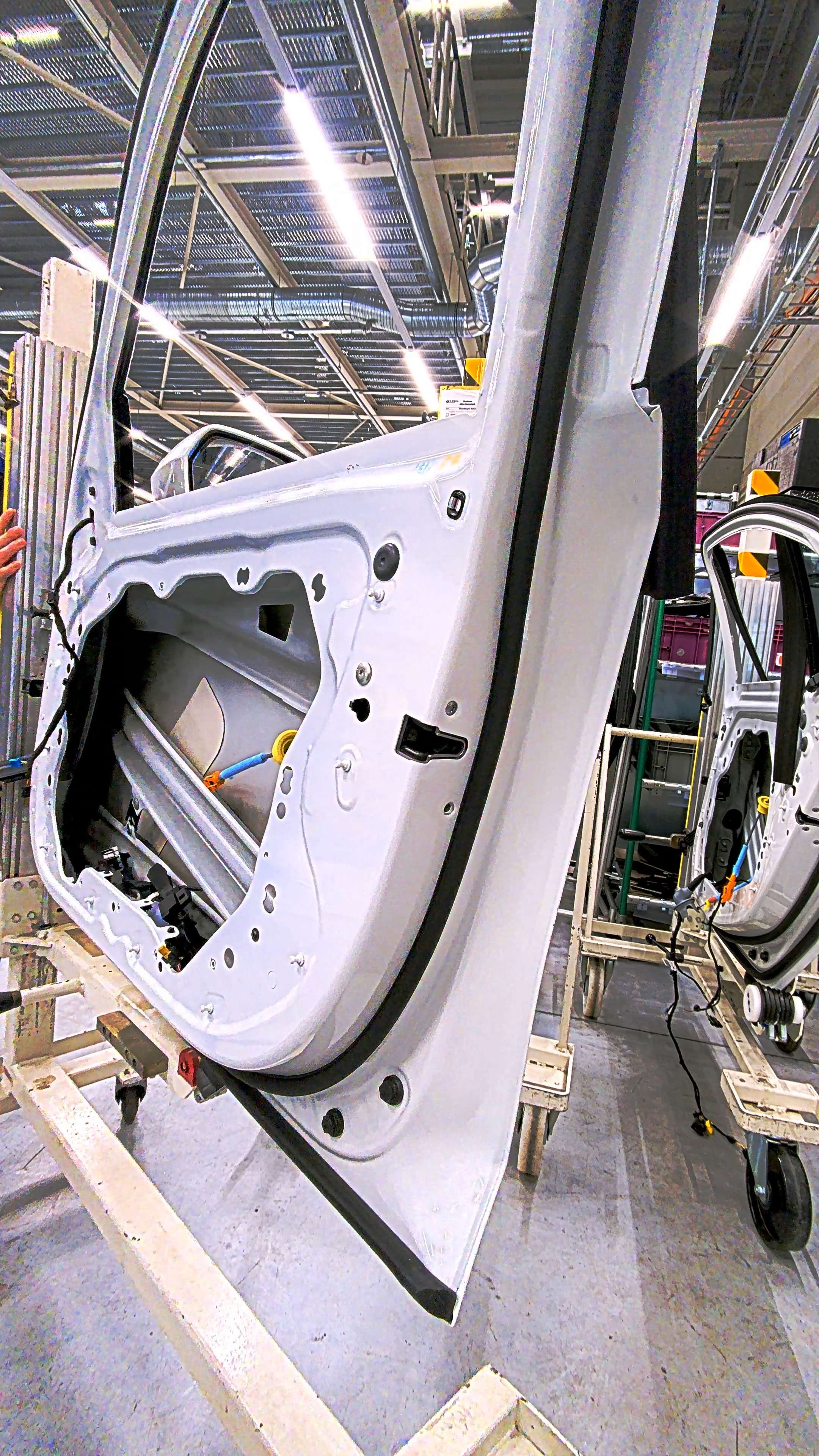}
  \end{subfigure}

   \caption{Some exemplary images from the ablation test set, taken by hand outside the production line. These images exhibit significant variance and distractions.
   }
   \label{fig:abl_data}
\end{figure}

During our initial experiments, it became apparent that the standard test set used in \Cref{ssec:poc} did not allow for a detailed study of performance influences. 
We created a new test dataset to verify the robustness of the model against certain variations that may occur in the production line.
This dataset was generated by manually capturing images outside the production line, allowing us to introduce a wide range of variations into the test images. We specifically focused on capturing images from different perspectives, with diverse backgrounds, and under varying lighting conditions. Additionally, we modified the geometry of the door by removing part of the interior and partially occluding the door with distractors.
A total of 300 images were captured. Some example images are shown in \Cref{fig:abl_data}.

\subsection{Object Material}

In this experiment, we compare our method of applying complex materials to the model with two prominent alternative solutions - photo-realistic textures \cite{akar_synthetic_2022, de_roovere_sim--real_2024} and random colors \cite{mayershofer_towards_2021, tobin_domain_2017-1}.
For the first option, we recreate the real scene as closely as possible. To introduce variance, we switch between real interior and exterior options.
In the latter approach, we apply basic, textureless materials and randomize the values for base color, roughness, specularity, and metalness.
We created datasets for both alternatives and tested them on images taken from the production line and on the ablation dataset. We compare the results to our approach and the benchmark trained on real images. The results are shown in \Cref{table:materials}. 

The models trained on photo-realistic textures and random colors failed to bridge the domain gap effectively. On the ablation dataset, all models exhibited poor performance, likely due to the high complexity of the images. However, our approach outperformed the others, achieving the highest accuracy on the dataset at 74.8\%, surpassing the others by at least 10\%. A possible explanation for these findings is that the use of complex materials in our approach provides significantly more variance, thus enabling better generalization across different domains.

\begin{table}
    \caption{The results for various material strategies on images captured on the production line and our ablation dataset. 
    }
    \label{table:materials}
    \begin{tabular*}{\hsize}{@{\extracolsep{\fill}}lcccccc@{}}
         \toprule
         & Dataset & A & P & R & mAP \\
        \colrule
        \multirow{2}{*}{Benchmark} & Production & 0.99 & 0.98 & 0.98 & 0.98 \\
        & Ablation & 0.50 & 0.63 & 0.61 & 0.65 \\
        \multirow{2}{*}{Complex Materials} & Production & 1.00 & 0.99 & 1.00 & 0.99 \\
        & Ablation & 0.74 & 0.92 & 0.81 & 0.92 \\
        \multirow{2}{*}{Photo-Realistic} & Production & 0.84 & 0.74 & 0.73 & 0.84 \\
        & Ablation & 0.40 & 0.57 & 0.71 & 0.67 \\
        \multirow{2}{*}{Random Color} & Production & 0.52 & 0.94 & 0.58 & 0.75 \\
        & Ablation & 0.64 & 0.87 & 0.73 & 0.86 \\
        \botrule
    \end{tabular*}
\end{table}

\subsection{Background}
Here, we evaluate the influence of the background scene on the model's performance. We compare seven different sets of backgrounds: real backgrounds, HDRIs, COCO \cite{fleet_microsoft_2014}, Describable Textures Dataset (DTD) \cite{cimpoi_describing_2014}, general factory scenes, several technical datasets from Kaggle \cite{kaggle}, and a mixture of all. 

\Cref{tab:bg} presents the results of our experiments. 
It shows a small performance difference between the models, which could be explained by training-to-training variances and the limited number of trainings per dataset.
Furthermore, it should be noted that all models achieved 100\% accuracy on the production line test set. We therefore argue that the background of the scene has negligible influence on the model's performance.

\begin{table}
  \caption{Influence of various backgrounds on performance, evaluated on the ablation dataset. 
  }
  \label{tab:bg}
  \begin{tabular*}{\hsize}{@{\extracolsep{\fill}}lcccc@{}}
    \toprule
     & A & P & R & mAP \\
    \colrule
    Real & 0.72 & 0.92 & 0.80 & 0.91  \\
    HDRI & 0.74 & 0.96 & 0.82 & 0.93  \\
    COCO & 0.73 & 0.95 & 0.81 & 0.92  \\
    DTD & 0.76 & 0.92 & 0.80 & 0.91  \\ 
    Factory & 0.74 & 0.96 & 0.81 & 0.92  \\
    Technical & 0.75 & 0.96 & 0.82 & 0.92  \\ 
    Mixed & 0.72  & 0.96 & 0.80 & 0.92  \\ 
    \botrule
  \end{tabular*}
\end{table}

\subsection{Noise and Distractors}
In the final experiment, we analyze the effects of added noise and distractors on synthetic images. We compare our approach to three datasets: one without Gaussian noise, one with primitive distractors (such as cubes and spheres), and one with complex 3D models from the YCB dataset \cite{calli_ycb_2015}. Random materials are applied to all distractors.

\Cref{tab:noise_distractors} shows that applying Gaussian noise results in an improved accuracy compared to training without noise, supporting Tobin et al.'s \cite{tobin_domain_2017-1} findings that Gaussian noise is beneficial in domain randomization. However, adding distractors reduces performance, which diverges from Tobin et al.'s findings that they are crucial for resilience to real-world distractions. This discrepancy might be due to partially occluded screws in the training dataset, which likely led to overfitting and decreased performance when encountering real distractors.

\begin{table}
\caption{Performance metrics for various noise and distraction strategies, evaluated on the ablation dataset.
}
\label{tab:noise_distractors}
\begin{tabular*}{\hsize}{@{\extracolsep{\fill}}lcccc@{}}
\toprule
 & A    & P    & R    & mAP \\ 
\colrule
Gaussian Noise & 0.74 & 0.92 & 0.81 & 0.92   \\
Without Noise & 0.69 & 0.94 & 0.79 & 0.90   \\ 
Primitive Distractors       & 0.64 & 0.96 & 0.72 & 0.88   \\ 
Complex Distractors         & 0.61 & 0.93 & 0.85 & 0.91   \\ 
\botrule
\end{tabular*}
\end{table}

\section{Discussion}
\label{sec:discussion}

In this section, we address some potential limitations of our study. A primary concern is our selection of only three use cases to demonstrate the applicability of synthetic data for visual quality inspection. Although this selection may appear limited, these cases are representative of a broad spectrum of real-world scenarios. Additionally, we have tested our approach on over ten other use cases not included in this paper, further validating the robustness and versatility of our method. 
Another limitation is our focus on static, rigid components, a deliberate choice due to their prevalence in automotive production lines. However, experiments have shown that inspections involving flexible components, such as wires, can be challenging because wires can be installed in a multitude of configurations. Manually modeling each of these configurations requires significant effort, and even then, the performance of the trained model is often underwhelming.
Furthermore, this method relies on the quality of the real images. In our experiments, the trained model exhibited poor performance on images that were heavily noisy, blurred, or poorly lit.
For future work, effective strategies for addressing such challenging applications should be developed.

\section{Conclusion}
\label{sec:conclusion}
In this work, we present an image generation pipeline for creating synthetic training data through domain randomization, aimed at training object detection models for visual quality inspection in automotive production. During rendering, we introduce extensive variation into the dataset by randomly applying complex materials to all components within the scene. We demonstrated in three real-world inspection scenarios, that models trained entirely on synthetic data can outperform those trained purely on real images.
We would like to emphasize the key advantages of utilizing this approach for synthetic images. The fully synthetic method enables training models without relying on real production images. This not only reduces development costs but also accelerates the data collection process by allowing the production of rare variants and defects on demand. Additionally, this method is simple to implement, making it accessible even to data scientists with minimal image generation experience.
Future work should explore the applicability of synthetic data across a broader range of complex use cases that involve, for example, non-rigid objects and challenging environmental influences.

\FloatBarrier



\end{document}